\documentclass[conference]{IEEEtran}
\IEEEoverridecommandlockouts
\usepackage{cite}
\usepackage{amsmath,amssymb,amsfonts}
\usepackage{algorithmic}
\usepackage{graphicx}
\usepackage{svg}
\usepackage{bm}
\usepackage{textcomp}
\usepackage{xcolor}
\usepackage{hyperref}
\usepackage{enumitem}
\usepackage{titlesec}
\def\BibTeX{{\rm B\kern-.05em{\sc i\kern-.025em b}\kern-.08em
    T\kern-.1667em\lower.7ex\hbox{E}\kern-.125emX}}

\titlespacing*{\subsection}
{0pt}{0.5ex plus 1ex minus .2ex}{0.5ex plus .2ex}
\begin{document}

% Macro for the name of the tool
%\def\keyword#1{\setformath{$\sf{#1}$}}
\def\keyword#1{$\sf{#1}$}
\newcommand{\toolname}{\keyword{RAGE}}

\newcommand{\eat}[1]{}

\title{
RAGE Against the Machine:
Retrieval-Augmented LLM Explanations
}

% Authors
\author{
\IEEEauthorblockN{Joel Rorseth}
\IEEEauthorblockA{
\textit{University of Waterloo}\\
jerorset@uwaterloo.ca}
\and
\IEEEauthorblockN{Parke Godfrey}
\IEEEauthorblockA{
\textit{York University}\\
godfrey@yorku.ca}
\and
\IEEEauthorblockN{Lukasz Golab}
\IEEEauthorblockA{
\textit{University of Waterloo}\\
lgolab@uwaterloo.ca}
\and
\IEEEauthorblockN{Divesh Srivastava}
\IEEEauthorblockA{
\textit{AT\&T Chief Data Office}\\
divesh@research.att.com}
\and
\IEEEauthorblockN{Jaroslaw Szlichta}
\IEEEauthorblockA{
\textit{York University}\\
szlichta@yorku.ca}
}

\maketitle

\begin{abstract}
This paper demonstrates RAGE, an interactive
tool for explaining Large Language Models (LLMs)
augmented with retrieval capabilities;
i.e., able to query external sources and pull
relevant information into their input context.
Our explanations are counterfactual in the sense
that they identify parts of the input context
that, when removed, change the answer to the
question posed to the LLM.
RAGE includes pruning methods to navigate
the vast space of possible explanations,
allowing users to view the provenance
of the produced answers.
\end{abstract}

\section{Introduction}

\textbf{Motivation.}
Artificial Intelligence (AI) has seen remarkable growth
in terms of both capability and popularity,
exemplified by recent large language models (LLMs)
such as OpenAI’s ChatGPT, Microsoft’s Copilot,
and Google’s Gemini.
The rapid progress in LLM capability is driven by scale,
as AI researchers train increasingly complex models
with increasingly large datasets
using enormous computational resources.
Within a short span,
state-of-the-art models have progressed
from training millions, to billions,
and now to trillions of internal parameters.
However, this increased complexity further obscures
the underlying decision-making process of LLMs,
making it challenging to rationalize or troubleshoot
their outputs. As LLMs are adopted in critical sectors,
it is imperative that verifiable explanations accompany
their outputs, to build trust.

The unique enhancements in capability
that distinguish LLMs from previous language models
amplify their explainability concerns.
Of particular relevance is
\textit{retrieval-augmented generation} (RAG),
a popular \textit{prompt engineering} strategy
that leverages a powerful new LLM capability
known as \textit{in-context learning}.
With RAG, an LLM augments its trained knowledge
by learning from external knowledge sources,
supplied directly via the LLM's input context (prompt).
RAG has been pivotal for LLMs
in reducing their tendency to hallucinate plausible
yet incorrect outputs. This complex process, however,
obfuscates the provenance of the produced answers.

\textbf{Background.}
Due to the recency of LLMs
and their emergent capabilities,
few efforts have been made to explain their phenomena.
Under the umbrella of \textit{mechanistic interpretability},
low-level analyses have sought to understand
the mechanisms behind transformer-based language models,
and capabilities like in-context learning,
by analyzing circuits that form amongst attention heads,
or by assessing an LLM's ability 
to override trained knowledge
\cite{wei2023larger}.
Our \textit{explainability} focus%
---%
which aims to trace the provenance of LLM answers during RAG%
---%
instead
demands high-level explanations of RAG in \textit{simple}
terms. While specific prompting strategies
such as \textit{chain-of-thought} (CoT) prompting
could serve as interpretable explanations,  
RAG has yet to receive dedicated focus
in the explainability literature.
RAG is a leading prompting strategy
for the use of modern LLMs
in question answering (QA),
as CoT and other prompting techniques are less applicable
and require specialized examples.
Specific concerns about RAG,
such as the lack of provenance in LLM answers,
or the ``lost in the middle'' context position bias
observed in recent LLMs
\cite{lostmiddle},
warrant dedicated study
under an explainability lens.

\textbf{Contributions.}
To fill this gap,
we demonstrate \toolname{},%
\footnote{%
    A video is available at
    \url{https://vimeo.com/877281038}.
}
an interactive tool designed to enable
\textit{RAG Explainability} for LLMs.%
\footnote{%
    The tool is available at
    \url{http://lg-research-2.uwaterloo.ca:8092/rage}.
}
Our tool deduces provenance and salience
for external knowledge sources used during RAG,
exposing the in-context learning behaviors of the LLM.
Motivated by our prior work
using counterfactual explanations
for information retrieval \cite{credence},
we derive provenance counterfactually
by identifying minimal context perturbations
that change an LLM's output.
Our contributions are summarized as follows.

\begin{enumerate}[nolistsep,leftmargin=*]
    \item \textbf{Answer Origin Explainability.}
    We introduce a novel framework
    to assess the origin of LLM answers,
    with respect to context knowledge sources,
    by evaluating counterfactual
    source combinations and permutations.
    
    \item \textbf{Pruning Strategies.}
    We present inference pruning strategies
    to reduce the space of possible counterfactual
    explanations, by prioritizing the evaluation
    of important context perturbations.
    
    \item \textbf{Interactive Demo.}
    Participants will pose questions
    to an LLM augmented with knowledge sources
    from real datasets. RAGE will display
    explanations for RAG scenarios
    where answers are ambiguous,
    sourced from inconsistent external knowledge,
    or traced through a chronological sequence.
    Participants will see how subjective questions,
    such as determining the greatest professional
    tennis player, can be answered differently by
    an LLM, depending on the combination
    and order of context sources.

\end{enumerate}

\section{System Description}

\subsection{Problem Description}

Open-book question answering is a task
where a system determines an answer
to a given question using common knowledge
about the topic and a provided set of sources.
In \toolname{}, we 
explain how an LLM performs this task,
using its own pre-trained knowledge
and retrieved knowledge sources.
A user initiates the process
by posing a search query $q$
to a retrieval model $M$.
Given an index of knowledge sources
and a relevance threshold $k$,
the retrieval model $M$ scores
and ranks the $k$ most relevant sources
from the index with respect to query $q$.
The resulting ordering of sources,
denoted as $D_q$,
forms a sequence we refer to
as the \textit{context}. 

We combine $D_q$ and $q$ to form
a natural language prompt $p$
for the LLM $L$. This prompt
instructs $L$ to answer question $q$
using the information contained
within the set of delimited sources
from $D_q$.
Although $p$ serves as the final and sole input
to the LLM,
we denote the answer $a$ produced
by the LLM for a given query $q$
and the sequence of knowledge sources $D_q$
as $a = L(q, D_q)$.
We also define $S(q, d, D_q)$
as the relative relevance score
of a source $d$ $\in$ $D_q$
with respect to the query $q$
and other sources within $D_q$.
To derive explanations,
we assess the answers generated
across various combinations or permutations
of the sources in $D_q$.
We refer in general to these two methods
as context \textit{perturbations}.

\subsection{Architecture}

\begin{figure}[t]
\centerline{
    \includegraphics[scale=0.4]{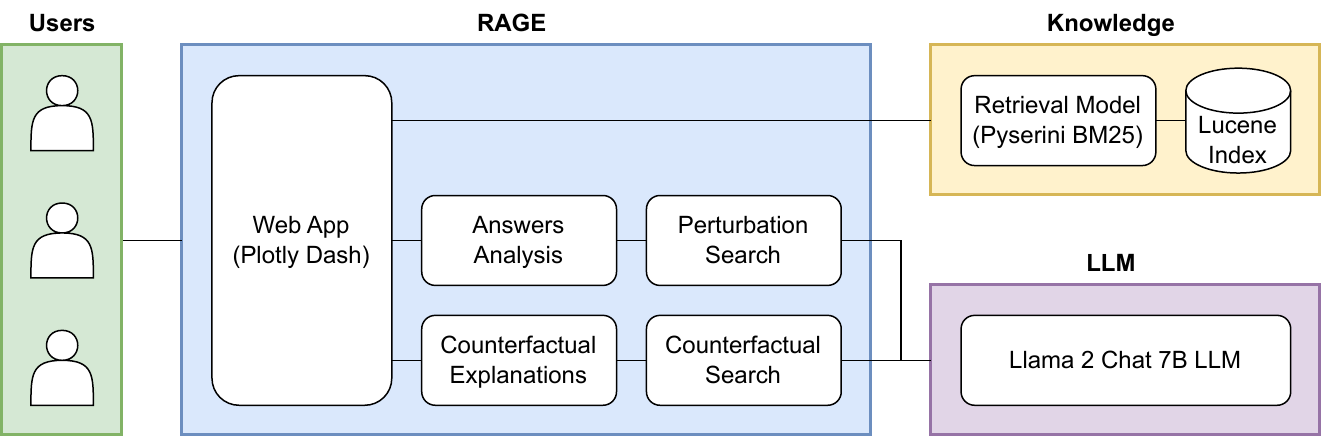}
}
\vskip -0.1in
\caption{The architecture of \toolname.}
\vskip -0.14in
\label{architecture}
\end{figure}

\toolname{} is an interactive Python web application developed
using the Plotly Dash web framework.
We installed the 7B Llama 2 Chat LLM
\cite{touvron2023llama}
(\textit{meta-llama/Llama-2-7b-chat-hf})
through the Hugging Face Transformers library.
Our software is,
however,
fully compatible with any similar transformer-based LLM.
All knowledge sources (documents) are retrieved
from our locally-configured document indexes,
using a BM25 retrieval model
from the Pyserini retrieval toolkit
\cite{pyserini}.

We run our application on an Ubuntu 22.04 server,
with an Intel Core i9-7920x CPU,
128GB of DDR4 RAM,
and an NVIDIA RTX 4090 GPU with 24GB of memory.
We use PyTorch's CUDA library
to run LLM operations on the GPU.

In \toolname{},
users can generate explanations
in terms of two complementary perturbations:
    source \textit{combinations}
    or
    source \textit{permutations}.
Combinations elucidate
how the \textit{presence}
of sources affects the LLM's predicted answer,
while permutations elucidate the effect
of their \textit{order}. Alongside
counterfactual explanations for each answer,
\toolname{} presents a pie chart
to visualize the distribution of answers,
a list of perturbation-answer rules,
and a table associating different answers
with the perturbations that produced them.

\subsection{RAG Explanations}

In generating counterfactuals, \toolname{} aims to
identify minimal perturbations to the context
that lead to a change in the LLM's predicted answer.
Combination-based counterfactual explanations,
which can serve as citations, may be generated
using a top-down or bottom-up search.
A top-down counterfactual must \textit{remove}
a combination of sources (subset of $D_q$) to flip
the \textit{full-context} answer to a target answer.
On the other hand, a bottom-up counterfactual
must \textit{retain} sources to flip the
\textit{empty-context} answer to the target answer.

In either case,
the candidate solution search space is defined
as the set of all combinations of the given sources.
We propose an iterative algorithm
that tests combinations in increasing order of
subset size. Specifically,
we evaluate all combinations containing $k$ sources
before moving on to those with $k+1$ sources.
Since there may be multiple combinations of equal size,
we iterate through these equal-size combinations
in order of their estimated relevance. 
This is calculated as the sum
of the relative relevance scores
of all sources within the combination,
which can be expressed as
$\sum_{d \in D_q} S(q, d, D_q)$.

To estimate the relative relevance
of a source $d \in D_q$,
the user can select
from two scoring methods $S$.
In the first method,
we aggregate the LLM's attention values,
summing them
over all internal layers,
attention heads,
and tokens
corresponding to a combination's constituent sources.
In the second method,
we sum the relevance scores produced
by the retrieval model for each source.
Since we only compare scores
for combinations of equal size,
there is no need to normalize combination scores
by the number of sources.

To generate permutation-based counterfactual explanations,
\toolname{} searches for the most similar source permutation
(with respect to their given order)
such that the LLM responds with a different answer.
These explanations quantify the stability of the
LLM's answer with respect to
the order of the context sources,
thus revealing any unexpected context position bias.
Our algorithm generates all length-$k$ permutations
for the $k$ sources,
then computes Kendall's Tau rank correlation coefficient
for each permutation
(with respect to their given order in $D_q$).
Once generated and measured,
the permutations are subsequently sorted and evaluated
in decreasing order of similarity,
based on decreasing Kendall's Tau.

For both combinations and permutations,
our algorithm continues until it finds
a perturbation that changes the answer,
or until a maximum number of perturbations
have been tested. Before comparing against
the original answer, we convert answers
to lowercase, remove punctuation, and
trim whitespace.

To supplement this counterfactual analysis,
we analyze the answers
over a selected set of perturbed sources.
To obtain a set of combinations,
\toolname{} considers all combinations
of the retrieved sources $D_q$,
or draws a fixed-size random sample
of $s$ combinations.
Based on the user's original question,
a prompt is created for each selected combination,
which is then used to retrieve
corresponding answers from the LLM.
After analyzing the answers,
\toolname{} renders a table
that groups combinations by answer,
along with a pie chart illustrating
the proportion of each answer
across all combinations.
A rule is determined
for each answer,
when applicable,
identifying sources
that appeared in all combinations
leading to this answer.

In a similar manner,
the user can instruct \toolname{}
to analyze answers from a selected set
of source permutations.
The table and pie chart illustrating associations
between answers and permutations
resemble those of the combination case,
with the rule calculation adopting a unique definition.
For each answer,
we determine a rule
that identifies any context positions
for which all permutations leading
to this answer shared the same source.
Users may again choose to analyze all permutations,
or a fixed-size random sample of $s$ permutations.

For the latter, a naive solution might generate
all $k!$ permutations of the $k$ sources,
then uniformly sample $s$ permutations,
resulting in $O(k!)$ time complexity.
To improve the efficiency,
we propose an implementation
using the Fisher–Yates shuffle algorithm
\cite{fisher1938statistical},
which produces an unbiased random permutation
of any finite sequence in $O(k)$.
In our approach,
we invoke the  Fisher-Yates algorithm $s$ times
to generate $s$ random permutations,
resulting in an efficient $O(ks)$ solution.

\toolname{} also allows the user to analyze 
the most \textit{optimum} permutations.
As observed in recent works \cite{lostmiddle},
LLMs often exhibit a context position bias,
paying more attention to sources appearing
at the beginning and end of the context
than those in the middle.
As a result,
sources that are important
for obtaining a given answer may not
receive sufficient consideration by the LLM.
Given a distribution
of the expected attention paid to each position,
this ``lost in the middle'' bias can be
counteracted by positioning important sources
in high-attention positions.
By requesting ``optimal permutations'' from \toolname{},
the user can analyze a set of permutations
with optimum placement of relevant sources
in high-attention positions.

To estimate the relevance of a source,
the user can choose to use
either the LLM's attention scores
or the retrieval model's assessed relevance
score. If desired,
the user can calibrate the expected distribution
of LLM context position attention by selecting
a predefined \textit{V-shaped} distribution.
Optimal permutations aim to maximize
both the relevance and attention
of their constituent sources.
A naive $O(k!)$ solution might generate
all $k!$ permutations,
scoring each by summing the product
of each source's relevance and attention,
then sorting and selecting
the $s$ highest-scoring permutations.
Recognizing that optimal permutations must
maximize both the relevance and attention
of their constituent sources,
we propose an efficient solution
by formulating this problem as an instance
of the \emph{assignment problem}
in combinatorics.

Numerous algorithms have been proposed
to solve this problem,
which aim to find the most optimal assignment
of all $k$ sources
to all $k$ context positions.
Since \toolname{} allows the user
to request the top-$s$ optimal permutations,
our formulation adopts a variant
of the assignment problem
that seeks the $s$ assignments
with minimal cost. We use the algorithm
proposed by Chegireddy and Hamacher
\cite{CHEGIREDDY1987155},
which allows us to calculate
the $s$ optimal permutations in $O(sk^3)$.

\begin{figure}[t]
\begin{minipage}[t]{0.485\textwidth}
  \includegraphics[width=\linewidth]{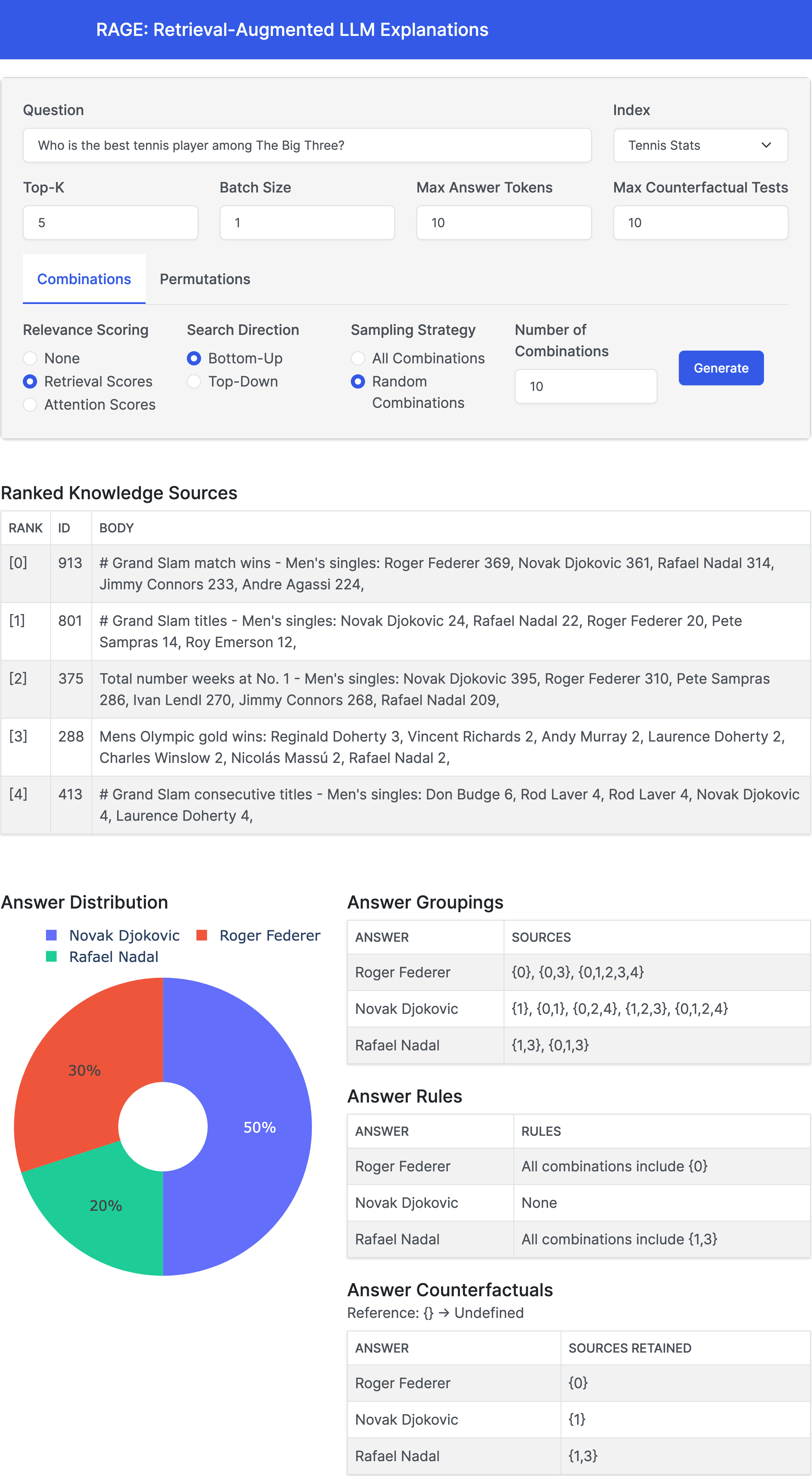}
  \vskip -0.1in
  \caption{
  Combination insights for the query
  about The Big Three.
  }
  \label{usecase1fig}
\end{minipage}
\vskip -0.14in
\end{figure}

\section{Demonstration Plan}

Conference participants will explore the
provenance of information included in
retrieval-augmented LLM responses.
They will then reinforce these findings
by evaluating the importance of relative
position among sources.

\subsection{Categorization of Use Cases}

The explanations generated by \toolname{} are
applicable across countless domains. Use cases
can be categorized based on various factors,
such as whether knowledge sources form a timeline,
or when questions are subjective,
leading to ambiguous answers.
In the former case,
\toolname{} identifies salient periods in time.
In the latter case,
it procures evidence to support various answers.
Knowledge sources may differ in terms of their consistency.
Our tool can identify consistent and inconsistent sources.
Sources may or may not share semantic dependencies,
and may or may not share syntactic formats.
\toolname{} will highlight source agreement and disagreement.

In the following subsections, we introduce
several use cases that highlight the
axes of this categorization. We begin by
exploring the possibility of an ambiguous answer,
which requires efficient evaluation over a large
sample of knowledge source combinations.
Next, we present a scenario
in which sources are slightly inconsistent,
testing \toolname's ability to identify minor
differences that can change the LLM's answer.
Last, we provide an example
in which the sources form a timeline,
requiring \toolname{} to strategically navigate
alternate timelines by minimally
combining and permuting the sources.

\subsection{Use Case \#1: Ambiguous Answers}

The user asks an LLM to determine
the best tennis player among ``The Big Three''
of Novak Djokovic, Roger Federer, and Rafael Nadal.
The user does not have any specific comparison
metric in mind, so they use the system to retrieve
a set of related documents,
each containing a different ranking
of The Big Three based on different metrics
(e.g., total number of match wins
and number of weeks ranked first).
The user expects that Novak Djokovic,
who recently surpassed Rafael Nadal and
Roger Federer in total Grand Slam wins,
might be the LLM's choice. But when asked
with the combination of all retrieved documents,
the LLM's answer is ``Roger Federer\@.''

Curious about why the LLM chose Federer,
the user poses the same query and documents
to \toolname, requesting combination insights.
As illustrated in Figure \ref{usecase1fig},
\toolname{} analyzes the answers generated
by the LLM using various combinations
of the given documents,
and discovers that the first document
led the LLM to produce this answer.
This document ranks various tennis players
based on total match wins,
with Federer ranking first at 369.
\toolname's answer rules formalize this explicitly,
asserting that this document was included in
every combination for which
the LLM answered ``Roger Federer.''

The user now comprehends
why the LLM chose Federer
but remains curious about the
document's relative significance.
Reviewing the original ranking,
they notice that this document has prominent placement
at the beginning of the context.
To investigate the impact of this position,
the user requests permutation-based explanations
for the same inputs.
Surprisingly,
\toolname{} reveals that moving
the document to the second position altered
the answer to ``Novak Djokovic\@.'' In short,
these explanations have enabled the user
to promptly identify the document
that influenced the LLM's answer,
and to understand the impact
of its relative position.

\subsection{Use Case \#2: Inconsistent Sources}

The user turns to an LLM
for help in determining the most recent winner
of the US Open women's tennis championship.
A small set of documents is retrieved,
each containing relevant statistics
about US Open championships.
The documents share similar format,
but some may be more current than others.
Hoping that the LLM will pinpoint
the most recent winner across all documents,
the user requests combination insights in \toolname{},
and observes how the combination
containing all sources produces the response
``Coco Gauff\@.'' With no further explanation,
the user aims to verify this result
by identifying the source document behind the answer,
and discovers that the last context document recognizes
Gauff as the 2023 champion.

Curious whether other out-of-date documents
could have been mistakenly sourced for an
incorrect answer, the user asks \toolname{}
to derive permutation insights.
By reordering the context documents
in various configurations and analyzing the
resulting answers, \toolname{} discovers
that the LLM incorrectly identifies
the 2022 champion ``Iga Swiatek''
whenever the last document is moved
towards the middle of the sequence. Using \toolname,
the user has  identified the up-to-date document
that offers the correct answer,
and has gleaned insights about out-of-date documents
and their ability to confuse the LLM.

\subsection{Use Case \#3: Timelines}

The user consults an LLM to determine
how many times Novak Djokovic won the
Tennis Player of the Year award between 2010 and 2019. 
The user gathers  relevant documents from the system,
each corresponding to one year's winner. Collectively,
the documents form a timeline for the three winners:
Rafael Nadal (2010, 2013, 2017, 2019),
Novak Djokovic (2011, 2012, 2014, 2015, 2018),
and Andy Murray (2016).

The user poses their question to \toolname,
which reports that the LLM produces
the expected answer of 5 when incorporating
the combination of all retrieved documents.
To validate the LLM's response,
the user expects an explanation listing each year
Djokovic won the award,
along with a citation to a supporting document.
To achieve this,
the user reviews the combination counterfactual
generated by \toolname{}
to determine the minimal set of documents
(and thus the exact years)
required to infer the correct answer.
\toolname{} cites five separate documents
from those provided,
each documenting a different year
in which Djokovic won Player of the Year.

Hoping to ensure that the LLM has not
overlooked any time period
covered by the documents,
the user asks \toolname{}
to derive permutation insights
over the same inputs.
By analyzing a sample of permutations,
the user is presented with a pie chart and answer table
that indicate a consistent answer of 5.
The user observed that no rules were found,
and concludes that the LLM consistently comprehends
the entire timeline of the twenty-tens,
regardless of the specific order
of the timeline's constituent documents.
Through \toolname,
the user has successfully discovered
which segments of the timeline were crucial
in determining the correct answer.

\bibliographystyle{IEEEtran}
\bibliography{IEEEabrv,references.bib}

\end{document}